\documentclass[smallabstract,smallcaptions]{dccpaper}

\usepackage{epsfig}
\usepackage{amsmath}
\usepackage{amssymb}
\usepackage{color}
\usepackage{url}
\usepackage{booktabs}
\usepackage{marvosym}
\usepackage{ifsym}
\usepackage[nameinlink,capitalise]{cleveref}

\newlength{\figurewidth}
\newlength{\smallfigurewidth}

\begin{document}

\title
{\large

\textbf{Butterfly: Multiple Reference Frames Feature Propagation Mechanism for Neural Video Compression}}

\author{%
Feng Wang$^{\ast}$, Haihang Ruan$^{\dag, \star}$, Fei Xiong$^{\ast}$, Jiayu Yang$^{\ast}$, \\ 
Litian Li$^{\ast}$ and Ronggang Wang$^{\ast}$\Letter\\[0.5em]
{\small\begin{minipage}{\linewidth}\begin{center}
\begin{tabular}{ccc}
$^{\ast}$Shenzhen Graduate School, Peking University, China\\ 
$^{\dag}$Shanghai Institute of Microsystem and Information Technology, \\
Chinese Academy of Sciences, Shanghai, China\\
$^{\star}$University of Chinese Academy of Sciences, Beijing, China\\
\url{wangfeng@stu.pku.edu.cn} \  \url{&} \ \url{rgwang@pkusz.edu.cn}\\
\thanks{\Letter: Corresponding contact author email: rgwang@pkusz.edu.cn.}
\small
\end{tabular}
\end{center}\end{minipage}}
}

\maketitle
\thispagestyle{empty}

\begin{abstract}
Using more reference frames can significantly improve the compression efficiency in neural video compression.
However, in low-latency scenarios, most existing neural video compression frameworks usually use the previous one frame as reference. 
Or a few frameworks which use the previous multiple frames as reference only adopt a simple multi-reference frames propagation mechanism.
In this paper, we present a more reasonable multi-reference frames propagation mechanism for neural video compression, 
called butterfly multi-reference frame propagation mechanism (Butterfly), which allows a more effective feature fusion of multi-reference frames.
By this, we can generate more accurate temporal context conditional prior for Contextual Coding Module. 
Besides, when the number of decoded frames does not meet the required number of reference frames, we duplicate the nearest reference frame to achieve the requirement, which is better than duplicating the furthest one.
Experiment results show that our method can significantly outperform the previous state-of-the-art (SOTA), and our neural codec can achieve -7.6$\%$ bitrate save on HEVC Class D dataset when compares with our base single-reference frame model with the same compression configuration.
\end{abstract}

\Section{1 \ \ Introduction}

Nowadays, video contributes to the majority of Internet traffic. Video data traffic on the Internet is growing faster than network bandwidth. So it is necessary to research and develop video compression technology to efficiently transmit the video data over the Internet with limited bandwidth. In the past few decades, many traditional video compression standards have been developed, such as H.264/AVC \cite{avc}, H.265/HEVC \cite{hevc}, H.266/VVC \cite{vvc}, etc. They usually use a hybrid coding framework, including intra/inter-frame prediction, transformer, quantization, entropy coding and loop-filter. However, these traditional video compression standards rely on hand-crafted modules. Although each module is well designed, they are difficult to optimize jointly.
Therefore, in order to obtain better compression efficiency, it is necessary to jointly optimize the video compression framework end-to-end.

\begin{figure*}
    \centering
     \includegraphics[width=0.75\linewidth]{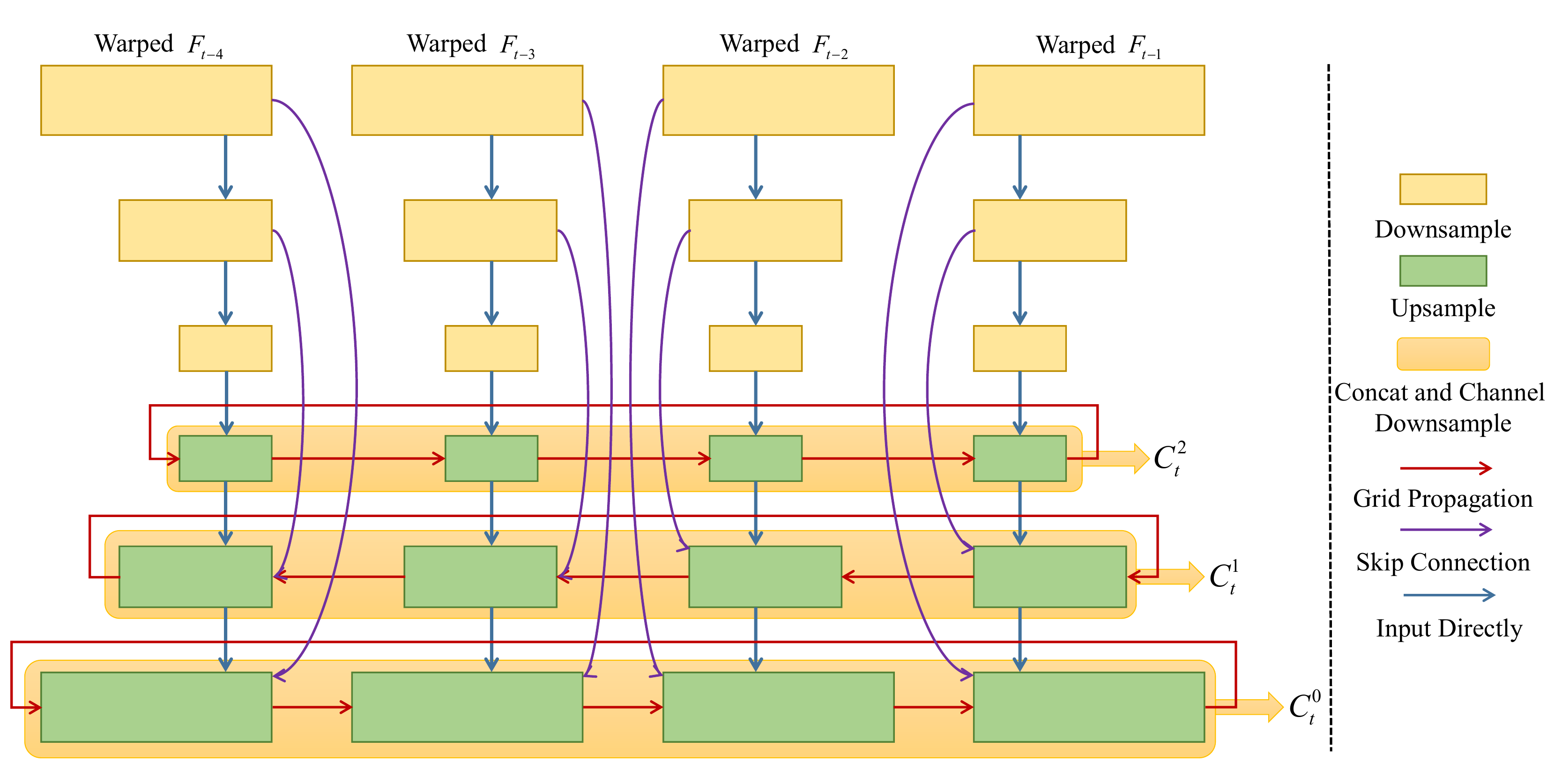}
     \vspace{-0.3cm}
     \caption{Structure of our proposed butterfly multi-reference frames feature propagation mechanism (Butterfly) module.}
     \vspace{-0.2cm}
     \label{fig:butterfly}
\end{figure*}
In recent years, with the rise and development of deep learning, more and more neural video compression (NVC) methods based on deep learning \cite{dcvc, tcm, mm2023, hlvc, rlvc, plvc, mlvc} have been proposed. 
According to the usage scenarios of video compression, 
existing NVC works can be divided into two categories. 
\footnote{Actually, the traditional VVC video compression standard contains three coding structures. They are All Intra mode (AI), Low Delay mode (LD) and Random Access mode (RA). However, in AI mode, each frame of the video sequence is encoded by intra coding (image compression), and the temporal dependency in the video sequence is not utilized. So the neural video compression method does not discuss the AI mode.} The first one is Low Delay (LD), 
and the other is Random Access (RA).
In this paper, we focus on the LD mode, 
which reduces the temporal redundancies by unidirectional prediction


In LD mode, there are a series of successful neural video codec \cite{ dcvc, tcm, mm2023, hlvc, rlvc, plvc, mlvc, dvcpro} have been designed, and they achieved very excellent compression efficiency. 
Most of them follow a similar process.
Firstly, the motion information between the previous frame and the current frame is obtained by using the motion estimation module, and then is compressed by the Motion Vector (MV) encoder-decoder. After obtaining the motion information, the motion compensation module performs motion compensation on the previous frame. And then using the compensation result as priori to use residual coding or conditional coding on the current frame.

However, most of them only use one previous frame as reference to predict the current frame. Some work \cite{rlvc, mlvc} use previous multiple reference frames 
and achieve better compression efficiency. But, their fusion methods for multiple reference frames are simple. For example, MLVC \cite{mlvc} only directly concats the four previous reference frames, and does not carry out a more targeted design. This will obliterate some key information in a frame when fusing these previous multiple reference frames. 

In this paper, we design a Butterfly Multi-Reference Frames Feature Propagation Mechanism (Butterfly) for Neural Video Compression (NVC) in Low Delay (LD) mode to get better compression efficiency. The Butterfly, as shown in \cref{fig:butterfly}, 
can fuse the features from different previous reference frames more effectively,
and will not cause some key information to be lost in the fusion process. 
So we can generate more accurate temporal context conditional prior for Contextual Coding Module and Frame Generator Module. 
Besides, when compressing the first \textbf{n-1} P-frames (assuming that the number of previous reference frames is \textbf{n}), whose reference frames are less than \textbf{n}, it is necessary to duplicate a certain reference frame to meet the required number of reference frames.
We prove by inequality that duplicating the nearest reference frame is better than duplicating the further one \cite{mlvc} in our model. 
Experiment results show that our method can significantly outperform the previous SOTA, and our method can achieve -7.6$\%$ bitrate save on HEVC Class D dataset when comprared with our base single-reference frame model \cite{tcm} with the same compression configuration.
The main contributions of this paper are summarized as follows:
\vspace{-1.0mm}
\begin{itemize}
\setlength{\itemsep}{0pt}
\setlength{\parsep}{0pt}
\setlength{\parskip}{0pt}
    \item We propose a Butterfly Multi-Reference Frames Feature Propagation Mechanism (Butterfly) for Neural Video Compression in LD mode. By using it, network can not only prevent some key information in a frame from being obliterated during fusing the multiple reference frames, but also aggregate information from different spatiotemporal locations and different scales.
    \item We prove by inequality that when decoded frames does not meet the required number of reference frames, it is better duplicating the nearest reference frame (Near-frame Duplication) .
    And we verify it by comparative experiment.
    \item Experimental results show that our method can significantly outperform the previous state-of-the-art (SOTA), and can achieve better compression efficiency comparing with our base single-reference frame model.
\end{itemize}

\Section{2 \ \ Butterfly Multi-Reference Frames Feature Propagation Mechanism}

We propose to utilize the butterfly multi-reference frames feature propagation mechanism to fuse the features of previous multiple reference frames. Then using the fusion result as prior to use conditional coding on the current frame.
In this section,
we firstly elaborate on our NVC framework in subsection 2.1, and then describe butterfly multi-reference frames feature propagation mechanism in subsection 2.2.

\SubSection{2.1 \ Overview}

\begin{figure*}
    \centering
     \includegraphics[width=0.75\linewidth]{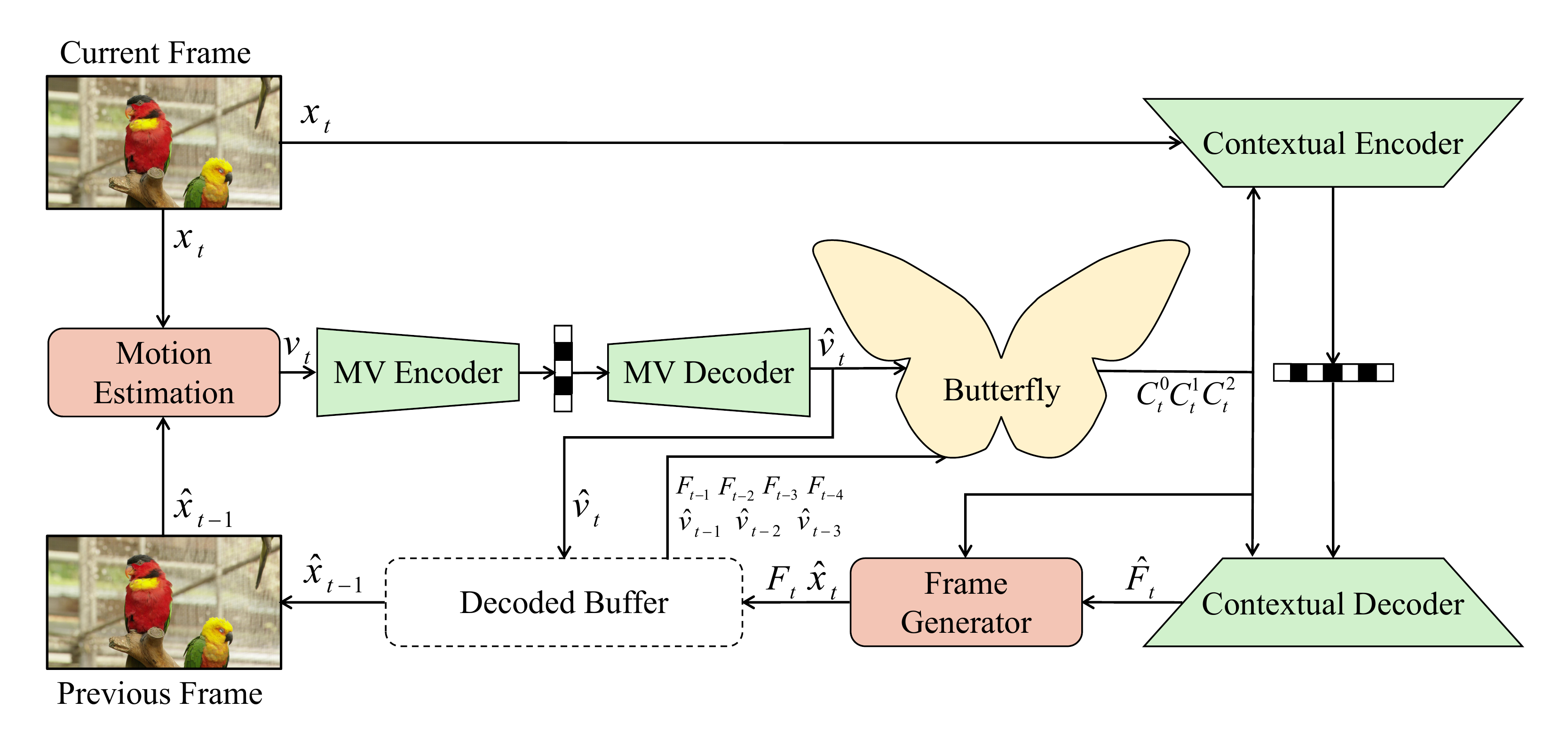}
     \vspace{-0.5cm}
     \caption{Overview of our proposed multi-reference frame neural video compression scheme. MV encoder-decoder and Contextual encoder-decoder all adopt hyper prior structure \cite{varia}.}
     \label{fig:overview}
     \vspace{-0.2cm}
\end{figure*}

Our network adopt the conditional coding-based framework of TCM \cite{tcm} and redesign reference frames feature fusion module. Fig. \ref{fig:overview} provides a high-level overview of our generalized neural video compression model, which contains the following five main sub-networks.

\textbf{Motion Estimation.} 
The correlation between adjacent frames is very strong in a video.
Temporal redundancy between frames can be reduced by motion estimation, which greatly reduces the number of bits of video transmission. In our proposed neural video compresison method, considering the computational complexity and model size of the entire model, we use the SPyNet \cite{spynet} to estimate the motion information. 

\textbf{MV Encoder-Decoder.} Transmitting the estimated motion information directly in the channel will consume a large number of bits, so it is necessary to compress the motion information. We utilize an auto-encoder style network with the hyperprior structure \cite{varia} as \cite{tcm}. 
The decoded $\hat{v}_{t}$ will be stored into the Decoded Buffer (DB).

\textbf{Butterfly Multi-Reference Frames Feature Propagation.} Most existing methods adopt the previous one frame as reference to compress the current frame. And seldom methods \cite{rlvc, mlvc} using previous multiple frames as reference only adopt a simple multi-reference frames propagation. These will limit the compression performance of neural video compression methods. 
In order to make full use of the information of multi-reference frames, we propose a Butterfly Multi-Reference Frames Feature Propagation Mechanism to fuse the information of multi-reference frames and generate conditional prior to guide the following Contextual Encoder-Decoder and Frame Generator to compress and reconstruct the current frame $x_t$. 
Specifically, in our network, warped multiple reference frames are passed into Butterfly to get multi-scale features.  
In the downsampling stage, multiple reference frames are independently downsampled to obtain features of different scales. In the upsampling stage, we use the grid propagation to fuse features of different scales.
By doing this, the network can not only prevent some key information in a frame from being obliterated during fusing the multiple reference frames, but also generate more accurate temporal context conditional prior by aggregating information from different spatiotemporal locations and different scales.
The details are presented in Subsection 2.2.

\textbf{Contextual Encoder-Decoder.} Following DCVC \cite{dcvc}, we adopt conditional coding-based method to compress the current frame $x_t$. 
For extracting more compact latent representation of $x_t$, we concat the temporal context conditional prior into Contextual Encoder-Decoder
for reducing temporal redundancy and spatial redundancy of $x_t$.
The output of Contextual Encoder-Decoder is the high-resolution feature $\hat{F}_t$ of current frame. We feed it into Frame Generator to generate reconstructed frame $\hat{x}_{t}$.

\textbf{Frame Generator.} It is used to generate the high-quality reconstructed frame $\hat{x}_{t}$. We use two plain residual blocks to build our frame generator. This can not only generate high-quality reconstructed frame, but also keep the complexity and parameters of the entire NVC model moderate.
The inputs of Frame Generator are the high-resolution feature $\hat{F}_t$ and the largest-scale temporal context conditional prior $\bar{C}^{0}_t$. 
The outputs are reconstructed frame $\hat{x}_{t}$ and feature $F_t$, which are stored into the Decoded Buffer (DB) to help compress the following frames.

\SubSection{2.2 \ Butterfly}

Using more reference frames can significantly improve the compression efficiency. 
However, simply concatting the information of different reference frames will lose some key information.
For example, due to occlusion, an object may only appear in one of the multi-reference frames. 
Concat simply concatnates the information of different frames in channel dimension (that is, the importance of each frame is same), then in the process of fusion, it will be eliminated by other frames.
Therefore, exploring a more effective multi-reference frames feature propagation mechanism can make full use of the temporal correlation information provided by multiple reference frames.

In order to fuse the features of multi-reference frames without losing key information in a certain frame. We design a butterfly multi-reference frames feature propagation mechanism, as shown in Fig. \ref{fig:butterfly}, which contains two main processes: upsampling and downsampling.  
In the downsampling stage, multiple reference frames are independently downsampled to obtain features of different scales. This ensures that some critical information will not be lost. In the upsampling stage, we use the grid propagation to fuse features of different scales and frames. This can aggregate information from different spatiotemporal locations and different scales and improve robustness of the network against occluded and fine regions.

Grid-like designs are seen in various vision tasks such as object detection, semantic segmentation, and video super-resolution \cite{basicvsr++}. By using this design in neural video compression, the module can adopt across resolutions to capture both fine and coarse information from different reference frames. Specifically, features of different scales are propagated between different previous reference frames in an alternating manner. Through propagation, the information from different frames can be “revisited” and adopted for feature fusion, 
thus improving feature expressiveness.

\Section{3 \ \ Near-Frame Duplication}

\begin{figure*}
    \centering
     \includegraphics[width=0.75\linewidth]{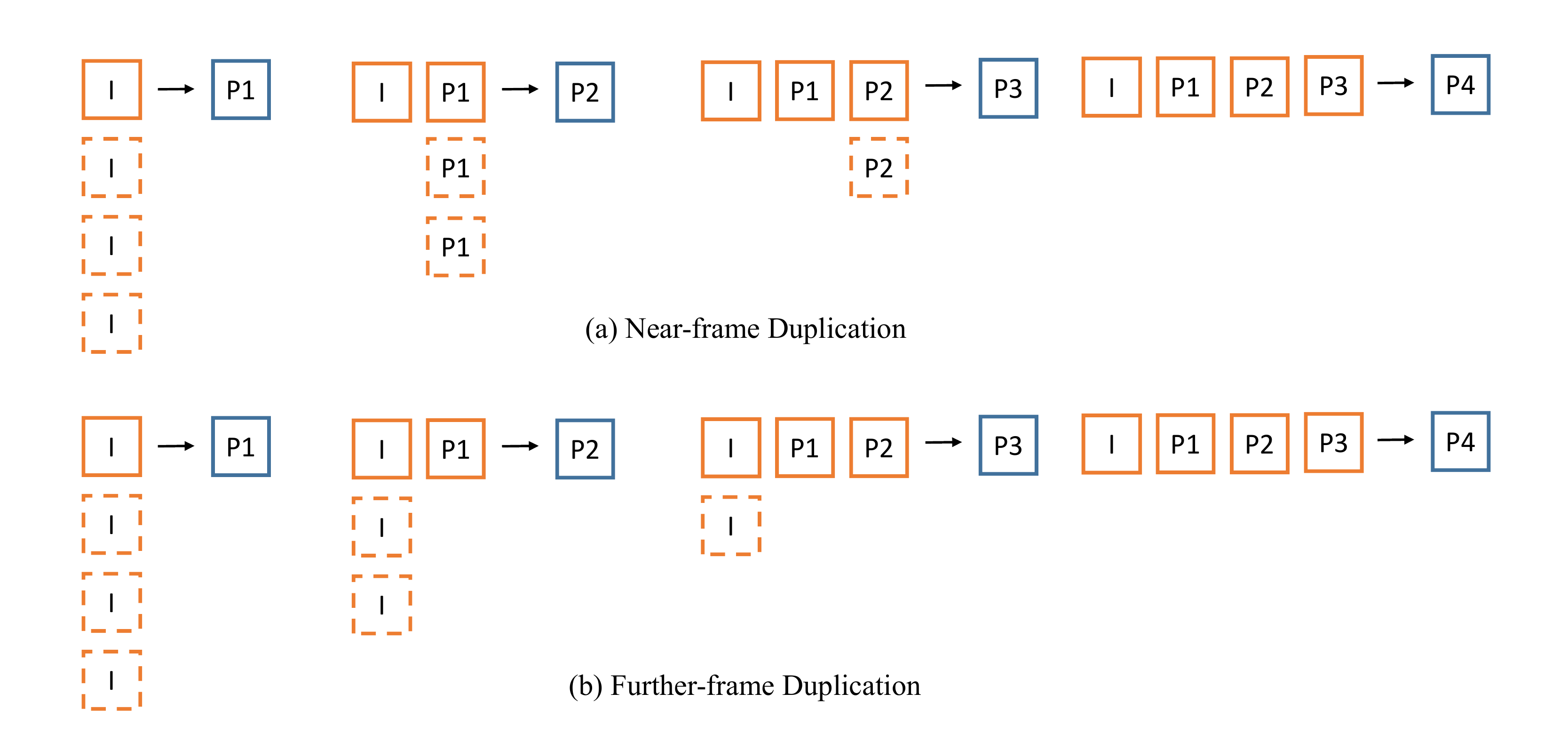}
     \vspace{-0.4cm}
     \caption{Methods of our proposed Near-Frame Duplication and MLVC \cite{mlvc} used Further-frame Duplication. The orange frame are reconstructed frames that have been decoded, the blue frame is the frame to be decoded, and the dotted frames represent duplicated frames.}
     \vspace{-0.2cm}
     \label{fig:nearduplicate}
  \end{figure*}

When decoded frames does not meet the required number of reference frames, it is necessary to duplicate a frame that has been decoded.
To achieve it, we can duplicate the nearest reference frame or the furthest one. 

We prove that when the number of reference frames can't meet the required number, it is better duplicating the nearest reference frame (Near-frame Duplication).
Since our neural video compression method is lossy compression, there is error accumulation in an 
intra period,
that is, the amount of information lost per frame in the temporal dimension increases gradually with compression process. For the convenience of analysis, we assume that the percentage increment of information lost in each frame compared to the previous frame is a fixed value $\boldsymbol{\alpha}$. That is
\begin{align}
    L_{t+1} = (1 + \boldsymbol{\alpha}) * L_t,
\end{align}
where $L_t$ and $L_{t+1}$ is the amount of information lost in frame at time step $t$ and $(t+1)$. 

Video is a sequence of frame images in the time axis direction, and the correlation between adjacent frames is very strong. For the convenience of analysis, we assume that the correlation between adjacent frames is also a fixed value $\boldsymbol{\beta}$. So we can get
\begin{align}
    I_{t+1} = \boldsymbol{\beta} * I_t,\ \  I_{t+2} = \boldsymbol{\beta}^2 * I_t
\end{align}
where $I_t$, $I_{t+1}$ and $I_{t+2}$ is the amount of information in frame at time step $t$, $(t+1)$ and $(t+2)$. 

Taking four reference frames as an example. For the convenience of analysis, we assume that each frame of the reference contributes same to the current frame.
When compress the first three P-frame, we duplicate the furthest reference frame. We have

\begin{small}
\begin{align}
    L_{1_f} &= L_1 \\
    L_{2_f} &= \frac{4 * L_{1_f} * (1 - \boldsymbol{\beta})}{4} * (1 + \boldsymbol{\alpha}) \\
    L_{3_f} &= \frac{3 * L_{1_f} * (1 - \boldsymbol{\beta}^2) + L_{2_f} * (1 - \boldsymbol{\beta})}{4} * (1 + \boldsymbol{\alpha}) \\
    L_{4_f} &= \frac{2 * L_{1_f} * (1 - \boldsymbol{\beta}^3) + L_{2_f} * (1 - \boldsymbol{\beta}^2) + L_{3_f} * (1 - \boldsymbol{\beta})}{4} * (1 + \boldsymbol{\alpha})
\end{align}
\end{small}
The total loss of information is
\begin{small}
\begin{equation}
\begin{aligned}
    \boldsymbol{L_{total_f}} &= L_{1_f} + L_{2_f} + L_{3_f} + L_{4_f}, \\
                            &= L_1 * (1 + (1 - \boldsymbol{\beta}) * (1 + \boldsymbol{\alpha}) + \frac{3 * (1 - \boldsymbol{\beta}^2) * (1 + \boldsymbol{\alpha})}{4} + \frac{(1 - \boldsymbol{\beta})^2 * (1 + \boldsymbol{\alpha})^2}{4} \\
                             &+ \frac{2 * (1 - \boldsymbol{\beta}^3) * (1 + \boldsymbol{\alpha})}{4} + \frac{(1 - \boldsymbol{\beta}) * (1 - \boldsymbol{\beta}^2) * (1 + \boldsymbol{\alpha})^2}{4} \\
                             &+ \frac{3 * (1 - \boldsymbol{\beta}) * (1 - \boldsymbol{\beta}^2) * (1 + \boldsymbol{\alpha})^2}{16} + \frac{(1 - \boldsymbol{\beta})^3 * (1 + \boldsymbol{\alpha})^3}{16}
                             )
\end{aligned}
\end{equation}
\end{small}

Similarly, when we duplicate the nearest reference frame. We have
\begin{small}
\begin{align}
    L_{1_n} &= L_1 \\
    L_{2_n} &= \frac{4 * L_{1_n} * (1 - \boldsymbol{\beta})}{4} * (1 + \boldsymbol{\alpha}) \\
    L_{3_n} &= \frac{L_{1_n} * (1 - \boldsymbol{\beta}^2) + 3 * L_{2_n} * (1 - \boldsymbol{\beta})}{4} * (1 + \boldsymbol{\alpha}) \\
    L_{4_n} &= \frac{L_{1_n} * (1 - \boldsymbol{\beta}^3) + L_{2_n} * (1 - \boldsymbol{\beta}^2) + 2 * L_{3_n} * (1 - \boldsymbol{\beta})}{4} * (1 + \boldsymbol{\alpha})
\end{align}
The total loss of information is
\begin{equation}
\begin{aligned}
    \boldsymbol{L_{total_n}} &= L_{1_n} + L_{2_n} + L_{3_n} + L_{4_n}, \\
                             &= L_1 * (1 + (1 - \boldsymbol{\beta}) * (1 + \boldsymbol{\alpha}) + \frac{(1 - \boldsymbol{\beta}^2) * (1 + \boldsymbol{\alpha})}{4} + \frac{3 * (1 - \boldsymbol{\beta})^2 * (1 + \boldsymbol{\alpha})^2}{4}\\
                             &+ \frac{(1 - \boldsymbol{\beta}^3) * (1 + \boldsymbol{\alpha})}{4} + \frac{(1 - \boldsymbol{\beta}) * (1 - \boldsymbol{\beta}^2) * (1 + \boldsymbol{\alpha})^2}{4} \\
                             &+ \frac{(1 - \boldsymbol{\beta}) * (1 - \boldsymbol{\beta}^2) * (1 + \boldsymbol{\alpha})^2}{8} + \frac{3 * (1 - \boldsymbol{\beta})^3 * (1 + \boldsymbol{\alpha})^3}{8} 
                             )
\end{aligned}
\end{equation}
\end{small}

Using Vieta Theorem and Differential Calculus of Single Variable
, we can get:
\begin{align}
    \boldsymbol{L_{total_n}} < \boldsymbol{L_{total_f}}
\end{align}
if 0 $<$ $\boldsymbol{\alpha}$ $<$ $\frac{-b + \sqrt{b^2 - 4 * a * c}}{2 * a} < 1$,
where $a = 5 * (1 - \boldsymbol{\beta})^3$, $b = - 9 * \boldsymbol{\beta}^3 + 17 * \boldsymbol{\beta}^2 - 27 * \boldsymbol{\beta} + 17$ and $c = 2 * \boldsymbol{\beta}^3 -28 * \boldsymbol{\beta}^2 - 30 * \boldsymbol{\beta}$. 

According to the concept of correlation, 0 $< \boldsymbol{\beta} <$ 1, so 0 $< \frac{-b + \sqrt{b^2 - 4 * a * c}}{2 * a} <$ 1. That means when the percentage increment of information lost $\boldsymbol{\alpha}$ less than a critical value in one NVC method, using Near-Frame Duplication can get smaller information loss and achieve better compression efficiency.

\Section{4 \ \ Experiments}

\SubSection{4.1 \ Experimental Setup}
We use the previous four decoded frames as reference following MLVC \cite{mlvc}.

\textbf{Datasets.} We use Vimeo-90k \cite{vimeo90k} for training, and HEVC Class B, C, D, E \cite{hevc}, HEVC Class RGB \cite{tcm}, UVG \cite{uvg} and MCL-JCV \cite{mcl-jcv} for testing.
When training, the Vimeo-90k videos will be randomly cropped into 256x256 patches.

\textbf{Implementation Detail.} We use the same image compression method as TCM \cite{tcm}. We optimized our model using quality metrics MSE (mean square error), and $\lambda$ is chosen from the set (256, 512, 1024, 2048). 
We use a step-by-step progressive training strategy to optimize the entire network. 
For the first 3 P-frames, we duplicate the nearest reference frame to achieve the required 4 frames.
We use the PSNR (peak single-to-noise ratio) 
to evaluate the distortion performance between the reconstructed frames and original frames. 
And we use the bpp (bit per pixel) to measure the number of bits for encoding the representations including MV and contextual latent code.
All the experiments are conducted on 2 NVIDIA V100 GPUs with a batch size of 4 and AdamW optimizer. 

\SubSection{4.2 \ Experimental Results}
We implement Butterfly on TCM \cite{tcm}, and use the same configuration as TCM for fair comparison. 
The intra period is set to 32 for the real applications.
To demonstrate the advantage of our proposed method, we compare with the existing traditional video codec including JM \cite{JM}, HM \cite{HM} and VTM \cite{VTM} 
\footnote{The software versions of JM, HM and VTM we use are JM-19.0, HM-16.2 and VTM-13.2, which represent the best encoder of H.264/AVC, H.265/HEVC and H.266/VVC respectively.} 
with the highest-compression-ratio settings for low-delay coding.
We also compare with previous NVC methods including DVCPro \cite{dvcpro}, MLVC \cite{mlvc}, RLVC \cite{rlvc}, DCVC \cite{dcvc} and TCM \cite{tcm}.

\begin{figure*}[t]
	\centering
	\vspace{-0.8cm}
	\includegraphics[width=0.9\linewidth]{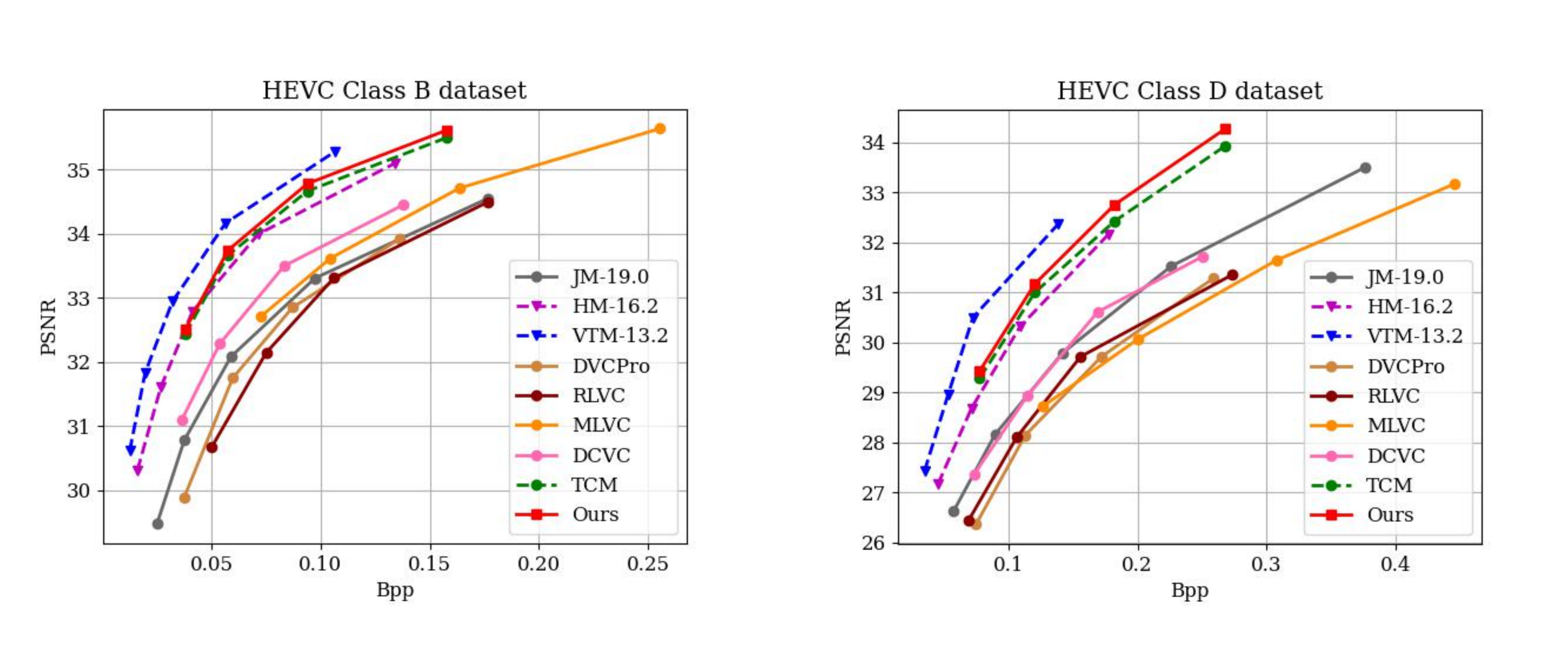} 
    \vspace{-1cm}
	\caption{PSNR and BPP curves.}
	\vspace{-0.8cm}
	\label{rd_curve}
\end{figure*}

Table \ref{tab_res_psnr} show the BD-Rate in different test datesets. We choose HM as anchor. Negative values indicate bit rate saving compared with HM while positive values indicate bit rate increasing. We can find that our method outperforms TCM \cite{tcm} which is our single-reference frame base mode on all test datasets. And it also surpasses the traditional codec HM on some datasets. Comparing with HM, our method achieve -12.6 $\%$ bitrate save on HEVC Class D dataset (means achieve -7.6$\%$ bitrate save compared with TCM).We also draw the RD-curves on HEVC Class B and D for intuitively presenting experimental results, as shown in Fig. \ref{rd_curve}.

\begin{table*}[t]
    \centering
    \caption{BD-Rate  (\%) comparison for PSNR. The anchor is HM-16.2.}
    \renewcommand{\arraystretch}{1.0}
    \small
      \begin{tabular}{ccccccccc}
      \toprule[1.0pt]
                              & B & C & D & E & RGB & UVG & MCL-JCV  \\ \hline
      HM-16.2  \cite{HM}      & 0.0    & 0.0     & 0.0    & 0.0    & 0.0    & 0.0    & 0.0      \\ \hline
      JM-19.0  \cite{JM}      & 96.9   & 56.6    & 50.0   & 80.5   & 102.4  & 108.1  & 95.4     \\ \hline
      VTM-13.2 \cite{VTM}     & -28.8  & -29.0   & -26.5  & -29.1  & -29.7  & -28.9  & -31.2    \\ \hline
      DVCPro   \cite{dvcpro}  & 123.7  & 124.0   & 93.6   & 283.0  & 102.1  & 137.7  & 99.3     \\ \hline
      RLVC     \cite{rlvc}    & 122.6  & 118.9   & 81.2   & 246.2  & 114.2  & 140.1  & 124.8    \\ \hline
      MLVC     \cite{mlvc}    & 61.4   & 124.1   & 96.1   & 138.8  & 82.1   & 66.5   & 66.8     \\ \hline
      DCVC     \cite{dcvc}    & 56.0   & 76.9    & 52.8   & 156.8  & 51.9   & 67.3   & 42.8     \\ \hline
      TCM      \cite{tcm}     & -5.3   & 15.1    & -5.4   & 18.5   & -14.4  & -9.0   & -3.2     \\ \hline
      Ours                    & -10.4  & 7.5     & -12.6  & 8.3    & -17.5  & -15.6  & -6.2    \\
      \bottomrule[1.0pt]
      \end{tabular}
      \vspace{-0.5cm}
    \label{tab_res_psnr}
\end{table*}

\SubSection{4.3 \ Ablation Study}
The ablation studies are conducted to verify the effectiveness of our proposed component (Butterfly) and method (Near-frame Duplication).
For simplification, all experiments use HEVC testsets. And the comparisons are measured by BD-Rate (\%).

\textbf{Butterfly.} 
In Butterfly, the downsampling process is independent, and the upsampling process fuses multi-scale features from different frames.
Therefore we design two comparative experiments.
The one is that upsampling and downsampling process are both together, which concats features before downsampling. 
This is the feature fusion method adopted by MLVC \cite{mlvc}.
The other is that upsampling and downsampling process are both independent, which concats features after upsampling.

\begin{table}[t]
    \vspace{-0.5cm}
    \caption{The method of Upsampling and Downsampling. }
    \vspace{-0.3cm}
    \centering
    \scalebox{0.8}{
    \renewcommand{\arraystretch}{1.25}
    \begin{tabular}{ccccccc}
    \toprule[0.9pt]
                                        & B       & C       & D        & E        & RGB        & Average \\ \hline
    Proposed Butterfly                  & 0.0     & 0.0     & 0.0      & 0.0      & 0.0        & 0.0     \\ \hline
    Up and Down Sampling Together       & 0.6     & 1.1     & 0.8      & 1.0      & 0.5        & 0.8     \\ \hline
    Up and Down Sampling Independent    & 0.8     & 1.2     & 0.7      & 1.0      & 0.8        & 0.9     \\ 
    \bottomrule[1.0pt]
    \end{tabular}}
     \vspace{-0.5cm}
    \label{butterfly ablation}
\end{table}

Table \ref{butterfly ablation} compares the performance of above three feature fusion methods. 
From this table, we can find that our method has the best performance of all.
This is because our method fully considers the independence of each frame's feature, and designs the feature fusion method in a targeted manner, rather than simply concat.

\textbf{Near-frame Duplication}
It is necessary to duplicate a frame that has been reconstructed to meet the requirement of four reference frames when compress the first three P-frame.
In Section 3, we prove that duplicating the nearest reference frame is better than duplicating the furthest one for our multi-reference frames NVC method.
In this subsection, we verify it by comparative experiments. 

\begin{table}[h]
    \vspace{-0.5cm}
    \caption{The position of the duplication frame. }
    \vspace{-0.3cm}
    \centering
    \scalebox{0.8}{
    \renewcommand{\arraystretch}{1.25}
    \begin{tabular}{ccccccc}
    \toprule[1.0pt]
                                          \ \    & B       & C       & D        & E        & RGB        & Average \\ \hline
    Near-frame Duplication (\textbf{Our}) \ \    & 0.0     & 0.0     & 0.0      & 0.0      & 0.0        & 0.0     \\ \hline
    Further-frame Duplication             \ \    & 1.4     & 1.0     & 1.5      & 2.2      & 1.6        & 1.5     \\ 
    \bottomrule[1.0pt]
    \end{tabular}}
     \vspace{-0.3cm}
    \label{mlvc duplication ablation}
\end{table}

We experiment Near-frame Duplication and Further-frame Duplication on Butterfly.
The comparative experiment results are shown in Table \ref{mlvc duplication ablation}.
From the table, we can find that Near-frame Duplication can bring significant performance gains.


\Section{5 \ \ Conclusion}
In this paper, we explore and design a butterfly multiple reference frames feature propagation mechanism for neural video compression. 
It can effectively fuse the features of multiple previous reference frames, prevent key information in a certain frame from being
eliminated during the fusion process, and generate more accurate temporal context conditional prior for following module. 
Besides, we prove that when the number of reference frames can't meet the required number, it is better duplicating the nearest reference frame.
Based on Butterfly and Near-frame Duplication, we propose a multi-reference frames neural video compression scheme for low-delay coding scenarios.
Experiment results show that our method can achieve higher compression efficiency compared with other existing neural video compression schemes.



\Section{Acknowledgment}
We thank Ren Yang (ETH Zurich) and Zongyu Guo (University of Science and Technology of China) for helpful discussions.
This work is supported by National Natural Science Foundation of China U21B2012 and  62072013, Shenzhen Cultivation of Excellent Scientific and Technological Innovation Talents RCJC20200714114435057, Shenzhen Research Projects of 201806080921419290.

\Section{References}
\bibliographystyle{IEEEbib}
\bibliography{refs}

\begin{thebibliography}{10}

\bibitem{avc}
Gary~J Sullivan, Pankaj~N Topiwala, and Ajay Luthra,
\newblock ``The h. 264/avc advanced video coding standard: Overview and
  introduction to the fidelity range extensions,''
\newblock {\em Applications of Digital Image Processing XXVII}, vol. 5558, pp.
  454--474, 2004.

\bibitem{hevc}
Gary~J Sullivan, Jens-Rainer Ohm, Woo-Jin Han, and Thomas Wiegand,
\newblock ``Overview of the high efficiency video coding (hevc) standard,''
\newblock {\em IEEE Transactions on circuits and systems for video technology},
  vol. 22, no. 12, pp. 1649--1668, 2012.

\bibitem{vvc}
Benjamin Bross, Ye-Kui Wang, Yan Ye, Shan Liu, Jianle Chen, Gary~J Sullivan,
  and Jens-Rainer Ohm,
\newblock ``Overview of the versatile video coding (vvc) standard and its
  applications,''
\newblock {\em IEEE TCSVT}, vol. 31, no. 10, pp. 3736--3764, 2021.

\bibitem{dcvc}
Jiahao Li, Bin Li, and Yan Lu,
\newblock ``Deep contextual video compression,''
\newblock {\em Advances in Neural Information Processing Systems}, vol. 34, pp.
  18114--18125, 2021.

\bibitem{tcm}
Xihua Sheng, Jiahao Li, Bin Li, Li~Li, Dong Liu, and Yan Lu,
\newblock ``Temporal context mining for learned video compression,''
\newblock {\em arXiv preprint arXiv:2111.13850}, 2021.

\bibitem{mm2023}
Jiahao Li, Bin Li, and Yan Lu,
\newblock ``Hybrid spatial-temporal entropy modelling for neural video
  compression,''
\newblock in {\em Proceedings of the 30th ACM International Conference on
  Multimedia}, 2022, pp. 1503--1511.

\bibitem{hlvc}
Ren Yang, Fabian Mentzer, Luc Van~Gool, and Radu Timofte,
\newblock ``Learning for video compression with hierarchical quality and
  recurrent enhancement,''
\newblock in {\em Proceedings of the IEEE/CVF Conference on Computer Vision and
  Pattern Recognition}, 2020.

\bibitem{rlvc}
Ren Yang, Fabian Mentzer, Luc Van~Gool, and Radu Timofte,
\newblock ``Learning for video compression with recurrent auto-encoder and
  recurrent probability model,''
\newblock {\em IEEE Journal of Selected Topics in Signal Processing}, vol. 15,
  no. 2, pp. 388--401, 2021.

\bibitem{plvc}
Ren Yang, Radu Timofte, and Luc Van~Gool,
\newblock ``Perceptual video compression with recurrent conditional gan,''
\newblock in {\em Processings of the IJCAI}, 2022.

\bibitem{mlvc}
Jianping Lin, Dong Liu, Houqiang Li, and Feng Wu,
\newblock ``M-lvc: Multiple frames prediction for learned video compression,''
\newblock in {\em Proceedings of the IEEE/CVF CVPR}, 2020.

\bibitem{dvcpro}
Guo Lu, Xiaoyun Zhang, Wanli Ouyang, Li~Chen, Zhiyong Gao, and Dong Xu,
\newblock ``An end-to-end learning framework for video compression,''
\newblock {\em IEEE TPAMI}, 2020.

\bibitem{varia}
Johannes Ball{\'e}, David Minnen, Saurabh Singh, Sung~Jin Hwang, and Nick
  Johnston,
\newblock ``Variational image compression with a scale hyperprior,''
\newblock {\em arXiv:1802.01436}, 2018.

\bibitem{spynet}
Anurag Ranjan and Michael~J Black,
\newblock ``Optical flow estimation using a spatial pyramid network,''
\newblock in {\em Proceedings of the IEEE CVPR}, 2017, pp. 4161--4170.

\bibitem{basicvsr++}
Kelvin~CK Chan, Shangchen Zhou, Xiangyu Xu, and Chen~Change Loy,
\newblock ``Basicvsr++: Improving video super-resolution with enhanced
  propagation and alignment,''
\newblock in {\em Proceedings of the IEEE/CVF Conference on CVPR}, 2022, pp.
  5972--5981.

\bibitem{vimeo90k}
Tianfan Xue, Baian Chen, Jiajun Wu, Donglai Wei, and William~T Freeman,
\newblock ``Video enhancement with task-oriented flow,''
\newblock {\em IJCV}, vol. 127, no. 8, pp. 1106--1125, 2019.

\bibitem{uvg}
Alexandre Mercat, Marko Viitanen, and Jarno Vanne,
\newblock ``Uvg dataset: 50/120fps 4k sequences for video codec analysis and
  development,''
\newblock in {\em Proceedings of the 11th ACM Multimedia Systems Conference},
  2020, pp. 297--302.

\bibitem{mcl-jcv}
Haiqiang Wang, Weihao Gan, Sudeng Hu, Joe~Yuchieh Lin, Lina Jin, Longguang
  Song, Ping Wang, Ioannis Katsavounidis, Anne Aaron, and C-C~Jay Kuo,
\newblock ``Mcl-jcv: a jnd-based h. 264/avc video quality assessment dataset,''
\newblock in {\em 2016 IEEE ICIP}. IEEE, 2016.

\bibitem{JM}
``{JM}-19.0,'' \url{http://iphome.hhi.de/suehring/},
\newblock Accessed: 2021-11-16.

\bibitem{HM}
``{HM}-16.20,'' \url{https://vcgit.hhi.fraunhofer.de/jvet/HM/-/tree/HM-16.20},
\newblock Accessed: 2021-11-16.

\bibitem{VTM}
``{VTM}-13.2,''
  \url{https://vcgit.hhi.fraunhofer.de/jvet/VVCSoftware_VTM/-/tree/VTM-13.2},
\newblock Accessed: 2021-11-16.

\end{thebibliography}

\end{document}